\documentclass[letterpaper, 10 pt, conference]{ieeeconf}
\IEEEoverridecommandlockouts                      
\usepackage{afterpage}
\usepackage{amsmath,amsfonts}
\usepackage{algorithmic}
\usepackage{algorithm}
\usepackage{array}
\usepackage{graphicx}
\usepackage[caption=false, font=small, labelfont=sf,textfont=sf]{subfig}
\usepackage{textcomp}
\usepackage{stfloats}
\usepackage{url}
\usepackage{verbatim}
\usepackage{graphicx}
\usepackage{cite}
\usepackage{xcolor}
\usepackage{soul}
\usepackage{svg}
\usepackage{multirow}
\usepackage[range-units=single,range-phrase=--]{siunitx}
\usepackage{soul}
\usepackage{fancyhdr}
\usepackage{hyperref}
\usepackage{booktabs}
\usepackage{svg}
\usepackage{tabularx}
\usepackage{acro}
\usepackage[table]{xcolor}

\definecolor{excelente}{RGB}{118,165,97}  % verde oscuro elegante
\definecolor{buena}{RGB}{198,224,180}     % verde claro
\definecolor{aceptable}{RGB}{255,235,156} % amarillo suave
\definecolor{deficiente}{RGB}{209,96,96} % rojo 
\newcommand{\newac}[2]{\DeclareAcronym{#1}{short=#1,long=#2}}

\newac{BASEPROD} {Bardenas Semi-Desert Planetary Rover Dataset}
\newac{ESA}     {European Space Agency}
\newac{ESTEC}   {European Space Technology and Research Center}
\newac{ExoTeR}  {Exomars Testing Rover}
\newac{FOG}     {Fiber Optic Gyro}
\newac{FPN}     {Feature Pyramid Network}
\newac{FoV}     {field of view}
\newac{GNC}     {Guidance{,} Navigation{,} and Control}
\newac{GNSS}    {Global Navigation Satellite System}
\newac{GPU}     {Graphics Processing Units}
\newac{HDPR}    {Heavy Duty Planetary Rover}
\newac{HR}      {High Resolution}
\newac{ICP}     {Iterative Closest Point}
\newac{IMU}     {Inertial Measurement Unit}
\newac{IoU}     {Intersection over Union}
\newac{JFR}     {Journal of Field Robotics}
\newac{LIBS}    {Laser Induced Breakdown Spectroscopy}
\newac{LRO}     {Lunar Reconnaissance Orbiter}
\newac{LTE}     {Long-Term Evolution}
\newac{LocCam}  {Localization Cameras}
\newac{MER}     {Mars Exploration Rover}
\newac{MRO}     {Mars Reconnaissance Orbiter}
\newac{MSL}     {Mars Science Laboratory}
\newac{MSR}     {Mars Sample Return}
\newac{MaRTA}   {Martian Rover Testbed for Autonomy}
\newac{NASA}    {National Aeronautics and Space Administration}
\newac{NeRF}    {Neural Radiance Fields}
\newac{NavCam}  {Navigation Cameras}
\newac{OBC}     {On Board Computer}
\newac{ORI}     {Ortho-Rectified Image}
\newac{PRL}     {Planetary Robotics Laboratory}
\newac{RAMS}    {Reliability{,} Availability{,} Maintainability{,} and Safety}
\newac{RAPID}   {Robust and (semi) Autonomous Platform for Increased Distances}
\newac{ROCC}    {Rover Operations Control Centre}
\newac{ROS}     {Rover Operating System}
\newac{RTK}     {Real Time Kinematic}
\newac{RoI}     {region of interest}
\newac{SfM}     {Structure-from-Motion}
\newac{SFR}     {Sample Fetching Rover}
\newac{SIFT}    {Scale-Invariant Feature Transform}
\newac{SI}      {International System of Units}
\newac{SLAM}    {Simultaneous Localization and Mapping}
\newac{SPA}     {Sense{,} Plan{,} Act}
\newac{SRL}     {Space Robotics Lab}
\newac{SR}      {Super Resolution}
\newac{SSL}     {Spectroscopy and Sensing Lab}
\newac{SVG}     {Scalable Vector Graphics}
\newac{UMA}     {University of Malaga}
\newac{UTM}     {Universal Transverse Mercator}
\newac{VO}      {Visual Odometry}
\newac{VSLAM}   {Visual Simultaneous Localization and Mapping}
\newac{PSNR}    {Peak Signal-to-Noise Ratio}
\newac{SSIM}    {Structural Similarity Index Measure}
\newac{LPIPS}   {Learned Perceptual Image Patch Similarity}

\DeclareAcronym{ROS2}{short=ROS~2,long=Robot Operating System 2}
\DeclareAcronym{LAENTIEC}{short=LAENTIEC,long=Area of Experimentation in New Technologies for Emergencies,foreign=\emph{Laboratorio y Area de Experimentación en Nuevas Tecnologías para la Intervención en Emergencias}}

\newcolumntype{Y}{>{\centering\arraybackslash}X}

\begin{document}

\title{High-fidelity 3D reconstruction for planetary exploration}

\author{A. Martínez-Petersen$^{1}$, L.~Gerdes$^{1}$, D. Rodríguez-Martínez$^{1}$, C.~J.~Pérez-del-Pulgar$^{1}$
\thanks{$^{1}$A. Martínez-Petersen, L.~Gerdes, C.~J.~Pérez-del-Pulgar, and D. Rodríguez-Martínez are with the Institute for Research in Mechatronics Engineering and Cyber-Physical Systems (IMECH) of the University of Málaga, Andalucía Tech, 29070 Málaga, Spain (e-mail: david.rm@uma.es).}}

\maketitle
\begin{abstract}
Planetary exploration increasingly relies on autonomous robotic systems capable of perceiving, interpreting, and reconstructing their surroundings in the absence of global positioning or real-time communication with Earth. Rovers operating on planetary surfaces must navigate under severe environmental constraints, limited visual redundancy, and communication delays, making onboard spatial awareness and visual localization key components for mission success. Traditional techniques based on \ac{SfM} or \ac{SLAM} provide geometric consistency but struggle to capture radiometric detail or to scale efficiently in unstructured, low-texture terrains typical of extraterrestrial environments. This work explores the integration of radiance field-based methods---specifically \ac{NeRF} and Gaussian Splatting---into a unified, automated environment reconstruction pipeline for planetary robotics. Our system combines the \textit{Nerfstudio} and \textit{COLMAP} frameworks with a \ac{ROS2}-compatible workflow capable of processing raw rover data directly from \textit{rosbag} recordings. This approach enables the generation of dense, photorealistic, and metrically consistent 3D representations from minimal visual input, supporting improved perception and planning for autonomous systems operating in planetary-like conditions. The resulting pipeline establishes a foundation for future research in radiance field–based mapping, bridging the gap between geometric and neural representations in planetary exploration.
\end{abstract}

\acresetall

\section{INTRODUCTION} 
\label{sec:introduction}

% - importancia de la robótica y la autonomía para la exploración espacial

Planetary exploration is rapidly evolving towards missions of increasing complexity and duration. In this context, rovers have become the main agents of observation and operation, adapting to increasingly extreme conditions over extended periods of time and reducing the costs and risks associated with direct human intervention in missions.

However, the challenges faced by rovers are numerous. Due to communication delays, absence of GNSS, and limited environmental knowledge, rover operators must rely on a combination of orbital maps, local images from the rover’s perspective, and the rover’s own pose estimation, which inherently contains errors. Given the possibility of losing a mission due to the failure to detect an obstacle, operations must be planned conservatively, making it essential to improve the autonomy and situational awareness of both the planning tools and the robot itself. 

To enhance such autonomy in the absence of reliable absolute references, current localization and perception systems in planetary exploration rely on multisensor data sources such as stereo cameras, \acp{IMU}, wheel encoders, laser rangefinders, and, in the near future, solid-state LiDAR sensors, to estimate relative motion and reconstruct the immediate environment. Techniques such as visual odometry allow the estimation of the trajectory and the generation of local maps by detecting and tracking features between successive stereo image pairs. However, the accuracy of these estimates can be severely limited under specific conditions: low-texture or homogeneous terrains (e.g., flat sand or uniform rock surfaces), large illumination changes (e.g., deep shadows or strong sunlight), low overlap between consecutive frames due to fast rover motion, and repetitive or ambiguous geometric patterns. Under these circumstances, visual correspondence becomes unreliable, causing pose errors to rapidly accumulate. Some proposed alternatives, such as \ac{SLAM}, face significant limitations in space missions due to their high energy consumption, sensitivity to harsh environmental conditions, and the low visual redundancy of rover paths, often resulting in sparse or noisy models that fail to capture the radiometric information of the environment adequately.

By contrast, recent 3D reconstruction approaches jointly integrate geometry, texture, and volumetric density into a single model, enabling the generation of high-fidelity, photorealistic reconstructions even from limited visual data. Unlike traditional geometric methods, continuous radiance-based representations are able to synthesize novel views, interpolate between camera positions, and capture complex lighting effects, opening new possibilities for visual perception and autonomous planning in planetary environments. Their direct use in robotic platforms remains constrained, however, by practical barriers, such as the need for extensive preprocessing, complex parameter tuning, and incompatibility with rover-specific data formats and sensor structures, which hinder their deployment beyond controlled research settings.

%Just to have a reference of the visual aspect and disposition of the page
\begin{figure*}[h!]
    \centering
    \includegraphics[width=\textwidth]{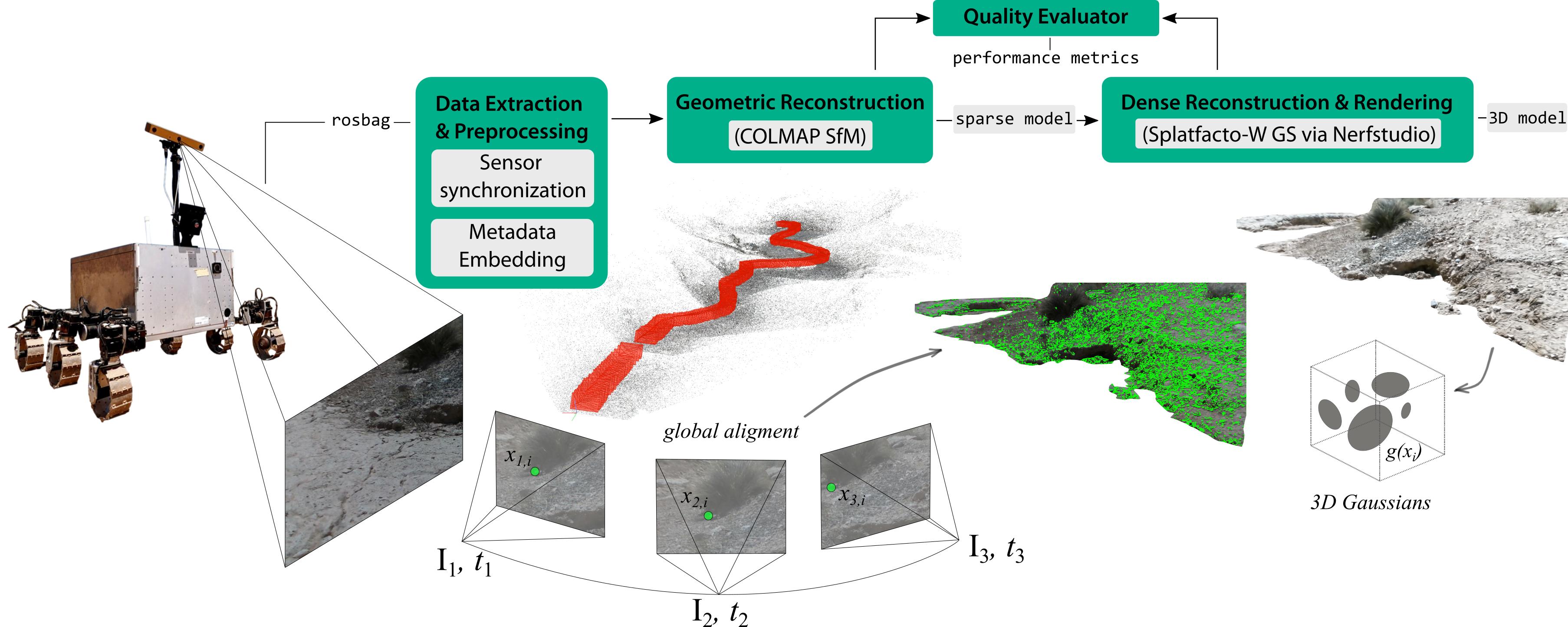}
    \caption{Overview of the proposed reconstruction pipeline. Raw rover data (such as RGB images, IMU readings, and GNSS data) are extracted directly from \textit{rosbag} recordings and preprocessed for geometric reconstruction with \textit{COLMAP} (\ac{SfM}). The resulting sparse model and camera poses are then passed to the radiance field stage, where \textit{Splatfacto-W} performs dense reconstruction through Gaussian Splatting, generating a photorealistic 3D model. Finally, a quality evaluation module computes performance metrics across both geometric and visual dimensions.}
    \label{fig:esquema}
\end{figure*}

% este es el párrafo más importante. si algo hay que depurar y que quede perfecto es este párrafo

The main contributions of this work are summarized as follows:

\begin{itemize}
    \item The design of a fully automated and modular pipeline for planetary environment reconstruction integrating \textit{Nerfstudio}~\cite{nerfstudio} and \textit{COLMAP}~\cite{colmap1,colmap2} within a unified workflow.
    \item Direct processing of raw ROS2 \textit{rosbag}~\cite{ros2} recordings, including multisensor synchronization, metadata enrichment with orientation priors, and automated format conversion for radiance field reconstruction.
    \item An enhanced geometric initialization strategy based on an externalized and updated \textit{COLMAP} \ac{SfM} stage, improving robustness and convergence under exploration-like conditions.
    \item A containerized implementation ensuring reproducibility, portability, and practical deployment across different computing platforms and operating systems.
\end{itemize}

\section{RADIANCE FIELD–BASED METHODS}

\subsection{Neural Radiance Fields (NeRF)}

NeRF are a 3D reconstruction technique that employs deep neural networks to continuously represent a three-dimensional scene from a set of 2D images~\cite{mildenhall2020nerf}. The model learns a function that maps a 3D position \(\mathbf{x}\) and a viewing direction \(\mathbf{d} = (\theta,\phi)\) (defined in spherical coordinates, where \(\theta\) is the polar angle and \(\phi\) the azimuthal angle) to a volumetric density \(\sigma(\mathbf{x})\) and color \(\mathbf{c}(\mathbf{x}, \mathbf{d}) = (R,G,B)\), as follows:

\begin{equation} 
\mathbf{C}(\mathbf{r}) = \int_{t_1}^{t_2} T \, \sigma \, \mathbf{c} \, dt
\end{equation}

where \(\mathbf{C}(\mathbf{r})\) is the expected color observed along camera ray \(\mathbf{r}(t) = \mathbf{o} + t \mathbf{d}\), and \(T(t)\) is the accumulated transmittance along the ray.

In its discrete form:

\begin{equation}
\mathbf{C}(\mathbf{r}) = \sum_{i=1}^{N} \alpha_i T_i \mathbf{c}_i, \quad
T_i = \exp\left(- \sum_{j=1}^{i-1} \sigma_j \delta_j \right),
\end{equation}

where \(i\) indexes the sampled points along each ray, \(\delta_i\) is the distance between consecutive samples, \(\alpha_i\) denotes their opacity and \(\mathbf{c}_i = (R_i,G_i,B_i)\) the corresponding color.

Essentially, NeRF is a model that learns how light and color exist in a scene. It uses a neural network to figure out what color each pixel should be by tracing rays of light from each camera pixel and learning the brightness and density of every point in space.

Several NeRF variants address similar challenges to those encountered in the 3D reconstruction of planetary environments. The original NeRF~\cite{mildenhall2020nerf} achieves very high-fidelity reconstructions but is computationally demanding. NeRF in the Wild (NeRF-W)~\cite{martinbrualla2021nerfwildneuralradiance} adds robustness to variable lighting and uncalibrated images, common in planetary exploration. Nerfacto \cite{nerfstudio} improves training efficiency, enabling the reconstruction of large scenes on resource-limited platforms. These variants exemplify the trade-offs between accuracy, robustness, and efficiency for rover trajectories.
% This implicit volumetric representation produces highly detailed results but requires considerable computational power and memory.

\subsection{Gaussian Splatting}

In contrast, the Gaussian Splatting method does not rely on neural networks but on an explicit representation based on 3D points associated with Gaussian distributions. Each 3D scene point \(\mathbf{x} \in \mathbb{R}^3\) is modeled as a Gaussian defined by its mean \(\boldsymbol{\mu}\), covariance matrix \(\Sigma\), and amplitude \(\alpha\):

\begin{equation}
G(x) = \alpha \exp\left[-\frac{1}{2} (x - \mu)^{T} \Sigma^{-1} (x - \mu)\right].
\end{equation}

During rendering, these 3D Gaussians are projected onto the 2D image plane according to the camera parameters. The projection is computed by transforming the covariance matrix from world coordinates to image space as:

\begin{equation}
\Sigma' = J W \Sigma W^{T} J^{T}, \quad \Sigma' \in \mathbb{R}^{2\times2}
\end{equation}

where \(W\) represents the world-to-camera transformation matrix, and \(J\) is the Jacobian of the perspective projection that maps 3D points onto the 2D image plane. This operation determines the apparent shape and spread of each Gaussian when viewed through the camera, producing the corresponding 2D \textit{splats} used in differentiable rendering. 

Unlike NeRF, which implicitly learns the scene through neural networks, Gaussian Splatting directly employs a set of 3D primitives (or \textit{Gaussians}), optimizing their position, color, size, and opacity parameters from multiple images. The result is a visually coherent reconstruction with lower computational cost and faster convergence times.  
While \ac{NeRF} produces highly detailed models at the expense of intensive training, 3D Gaussian Splatting achieves a more efficient and explicitly parameterized representation that is easier to update, making it particularly suitable for time-constrained or resource-limited applications~\cite{kerbl2023gaussian}. Variants of the original method as Splatfacto ~\cite{kerbl2023gaussian}, improve Gaussian Splatting by jointly optimizing the Gaussians for both efficiency and reconstruction quality. Its extension, Splatfacto in the Wild (Splatfacto-W)~\cite{xu2024splatfactow}, further improves robustness to varying lighting conditions and uncalibrated camera data, similar in spirit to NeRF-W. 

\vspace{5 mm}
\begin{table*}[h!]
\centering
\caption{Comparison of NeRF and Gaussian Splatting variants relevant to rover-based 3D reconstruction.}
\small
\begin{tabularx}{\textwidth}{lXXXXXX}
    \toprule
    \textbf{Parameter} 
    & \textbf{NeRF} 
    & \textbf{NeRF-W} 
    & \textbf{Nerfacto} 
    & \textbf{3DGS} 
    & \textbf{Splatfacto} 
    & \textbf{Splatfacto-W} \\
    \midrule
    Processing time & Slow & Slow & Medium & Fast & Very fast & Fast \\
    Reconstruction accuracy & Very high & High & High & High & High & Medium–High \\
    Memory consumption & Very high & Very high & High & High & Medium & Medium \\
    Robustness to lighting / calibration & Low & High & Medium & High & High & Very High \\
    Suitability for rover trajectories & Low & Medium & Medium & Medium & High & Very High \\
    \bottomrule
\end{tabularx}

\label{tab:nerf_splatting_comparison}
\end{table*}

\subsection{Comparison of Methods Applied to Space Robotics}

Selecting the most suitable radiance field approach for planetary rover data requires balancing visual fidelity, computational efficiency, and robustness to challenging acquisition conditions. The comparison presented in \autoref{tab:nerf_splatting_comparison}  does not aim to be exhaustive but rather to justify the selection of the most appropriate model for our proposed pipeline. Six representative variants were selected for their relevance to planetary exploration scenarios, which are characterized by irregular trajectories, limited image overlap, variable illumination, and constrained onboard computing resources. The parameters reported in \autoref{tab:nerf_splatting_comparison} capture complementary aspects of performance: while processing time, GPU memory consumption, and reconstruction accuracy describe the expected trade-offs between efficiency and quality, robustness to lighting and calibration reflects how well each method tolerates real-world imaging imperfections. Finally, the suitability for rover trajectories highlights each method’s capacity to handle the motion patterns typical of planetary rovers, such as long straight segments, low parallax, and sudden viewpoint changes, which challenge conventional radiance field reconstruction pipelines.

Within NeRF-based methods, the original NeRF provides the reference for maximum visual quality, NeRF-W introduces robustness to uncalibrated and uncontrolled data, and Nerfacto optimizes the trade-off between performance and reconstruction accuracy. On the other hand, Gaussian-based methods such as 3D Gaussian Splatting, Splatfacto, and Splatfacto-W offer explicit, faster, and more flexible representations better suited for real-world and resource-limited applications.

Given the characteristics of rover datasets---raw RGB images, variable-velocity rover motion, and strong illumination variability---\textit{Splatfacto-W} was identified as the most appropriate method. This variant combines the computational efficiency of Gaussian Splatting with the robustness mechanisms of NeRF-W, enabling stable reconstruction under uncontrolled conditions while maintaining low memory requirements and high adaptability to non-uniform trajectories. Consequently, it was selected as the reconstruction core of the proposed pipeline.

\section{METHODOLOGY}
\label{sec:methodology}

\subsection{Pipeline Description}

The proposed pipeline automates the complete 3D reconstruction workflow, from raw rover data to fully rendered visual models through a series of modular terminal scripts that ensure full reproducibility, scalability, and minimal user intervention. Each stage is independently executable yet seamlessly integrated into a unified framework, standardizing parameter configurations and guaranteeing consistent performance across experiments and hardware setups.

The pipeline, as described in \autoref{fig:esquema}, is organized into four main stages, each corresponding to a key phase in the results analysis:

\begin{enumerate}
    \item \textbf{Data Extraction and Preprocessing:} Raw mission data are extracted directly from \ac{ROS2} \textit{rosbags}. This stage automates the selection of RGB image sequences, the import of intrinsic and extrinsic camera parameters, and the registration of GNSS or odometry-based coordinates when available. The resulting dataset is formatted for geometric reconstruction and can optionally include localization priors to guide subsequent stages.
    
    \item \textbf{Geometric Reconstruction (\textit{Structure-from-Motion}):} Unlike the default \textit{Nerfstudio} integration, this pipeline employs an external instance of \textit{COLMAP}, configured with extended functionality for pose initialization and prior-based optimization. Different \textit{preprocessing configurations} can be applied---such as \textit{sequential matching}, which restricts feature matching to temporally adjacent frames, \textit{exhaustive matching}, which tests all image pairs, or the proposed hybrid method combining sequential matching with camera priors---to optimize correspondence search and convergence stability. The output includes camera trajectories, reconstructed 3D points, and geometric quality metrics such as reprojection error, number of feature observations, and track length.
    
    \item \textbf{Dense Reconstruction and Rendering:} The output of \textit{COLMAP} is automatically converted into the format required by \textit{Nerfstudio}, where dense 3D reconstruction is performed based on \textit{Splatfacto-W}. This step produces high-fidelity photorealistic models that can be interactively visualized and rendered from novel viewpoints, allowing both qualitative inspection and quantitative evaluation of visual consistency.
    
    \item \textbf{Quantitative and Qualitative Evaluation:} The final stage extracts and aggregates performance metrics to assess both geometric and visual quality. Geometric consistency is quantified through reprojection error, number of feature observations per image, and mean track length, while rendering fidelity is evaluated using perceptual image metrics such as \ac{PSNR}, \ac{SSIM}, and \ac{LPIPS}. These indicators provide a comprehensive view of reconstruction accuracy, radiometric realism, and overall system robustness.
\end{enumerate}

All stages are orchestrated within a unified execution environment to ensure reproducibility and efficiency. To enhance portability, the complete pipeline is containerized using Docker, integrating dedicated environments for \textit{COLMAP}, \textit{Nerfstudio}, and \ac{ROS2}, together with all dependencies. This design allows seamless deployment across different operating systems and computational infrastructures, including local workstations and GPU clusters. The modular structure also facilitates the integration of alternative reconstruction backends or sensor modalities, extending the framework’s applicability to different mission scenarios.

\subsection{Experimental Setup}
The pipeline was executed on a high-performance server, equipped with multiple NVIDIA~A6000 GPUs, high-speed NVMe storage, and a \qty{10}{Gbps} NAS system, enabling real-time visualization and the reconstruction of large-scale datasets. This infrastructure ensures efficient data management, high throughput, and the capability to process extended rover trajectories that exceed the memory limits of standard desktop systems.

\subsection{Pipeline Validation}

The proposed workflow was validated using real rover data from the \ac{BASEPROD} dataset~\cite{Gerdes2024Baseprod}, acquired by the \ac{PRL} of the \acf{ESA} and the \ac{SRL} of the \acf{UMA} using the \ac{MaRTA} rover~\cite{MartaExoter}. The experiments were carried out in the Bardenas Reales semi-desert (Navarra, Spain), a Mars-analog site characterized by heterogeneous textures, irregular terrain geometry, and highly variable natural illumination. Although the data were not originally captured for photogrammetric purposes—resulting in motion blur and non-optimized viewpoints—these conditions make the dataset a realistic testbed for assessing reconstruction performance under rover-like constraints. The full dataset, including both the extracted data and the original \ac{ROS2} \textit{rosbags}, is publicly available in ESA's robotics datasets repository~\cite{BaseprodData}.

Five representative segments were selected from different traverses to ensure diversity in visual appearance, trajectory type, and image count.
Trajectories T1 through T5 in this paper correspond to segments of traverses 2023-07-22\_14-18-23, 2023-07-21\_12-38-15, 2023-07-23\_13-05-11, 2023-07-23\_12-52-39, and 2023-07-21\_17-34-18 in \ac{BASEPROD}.
As we needed a different format than the already extracted data for this application, we extracted the required data directly from the \textit{rosbags} using the automated data extraction stage of the pipeline.  

Each segment was processed independently through the complete workflow, comprising data extraction, geometric reconstruction with \textit{COLMAP}, and dense visual modeling using \textit{Splatfacto-W}. 

The experiments were conducted under outdoor lighting with variable solar incidence angles and included slight deviations in rover speed and camera orientation. This configuration provides a realistic approximation of planetary rover operations, testing the robustness of the proposed method against real-world noise, imperfect calibration, and heterogeneous imaging conditions.

\section{RESULTS}
\label{sec:results}

This section summarizes the outcomes of applying the proposed pipeline to the rover datasets from the \ac{BASEPROD} dataset, as described in \autoref{sec:methodology}. 
Both qualitative and quantitative results are presented to assess reconstruction accuracy, visual fidelity, and computational efficiency.

\subsection{Overview of the Reconstructions}

\begin{figure*}[t]
    \centering
    % --- Primer bloque: vistas aéreas lado a lado ---
    \begin{minipage}[t]{0.48\textwidth}
        \centering
        \includegraphics[clip, trim={0cm 1.15cm 0cm 3cm}, width=\textwidth]{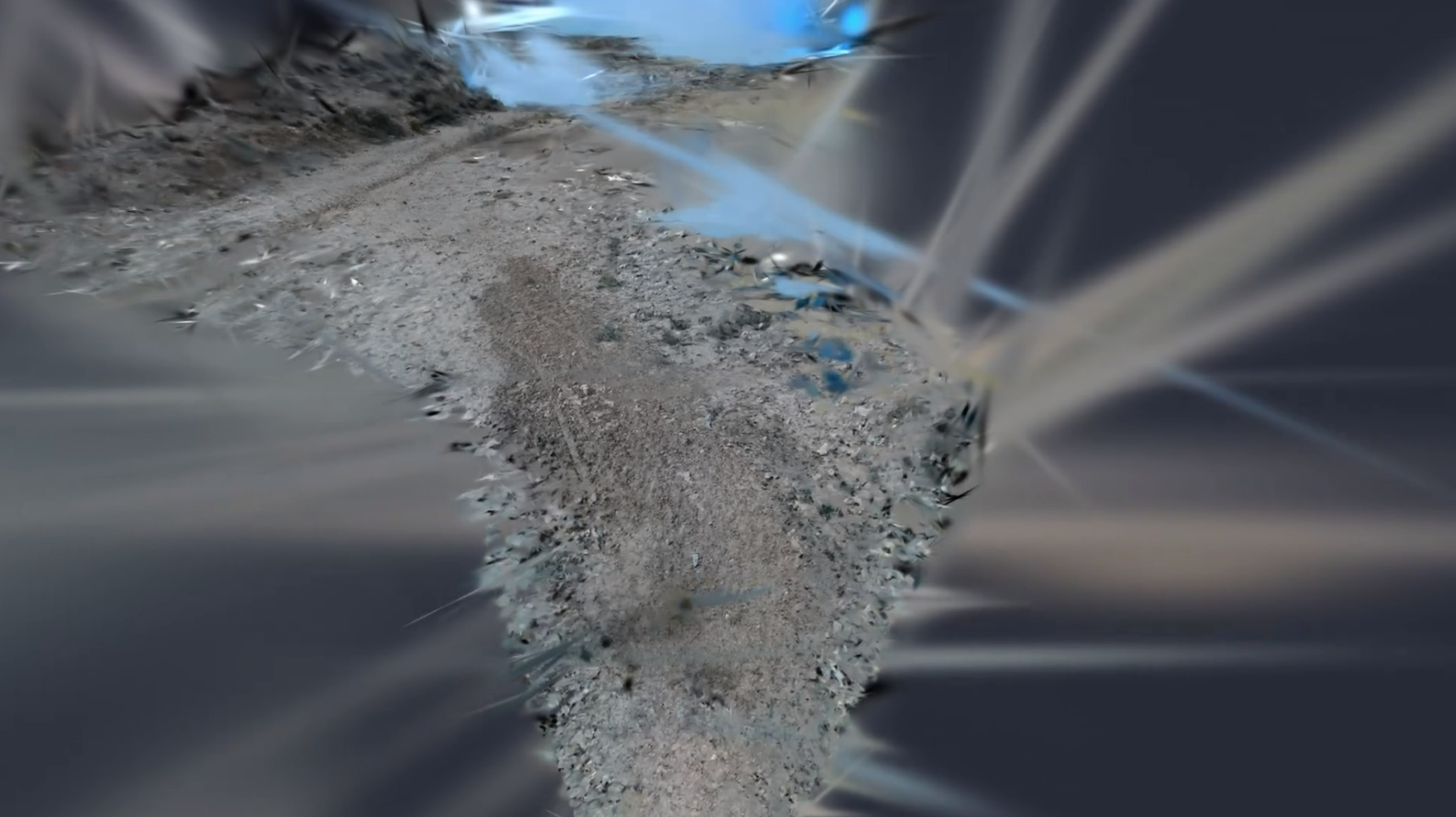}
    \end{minipage}\hfill
    \begin{minipage}[t]{0.48\textwidth}
        \centering
        \includegraphics[width=\textwidth]{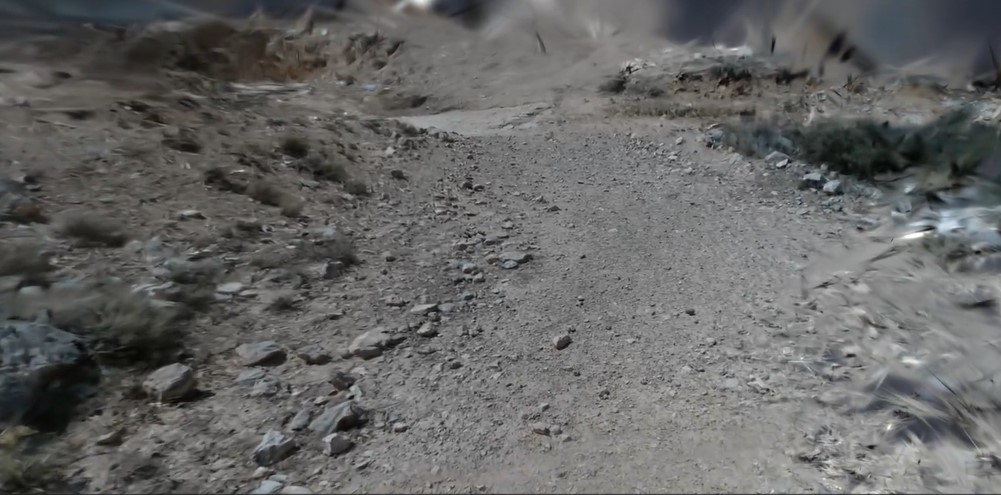}
    \end{minipage}
    \vspace{2mm}
    \caption{General aerial views of 3D models reconstructed from rover traverses in \ac{BASEPROD}. Each model corresponds to a segment of the terrain captured by the MaRTA rover under real outdoor illumination and surface conditions.}
    \label{fig:aerial_views}
\end{figure*}

\begin{figure*}[t]
    \centering
    % --- Segundo bloque: imágenes alargadas a ancho completo ---
    \includegraphics[width=\textwidth]{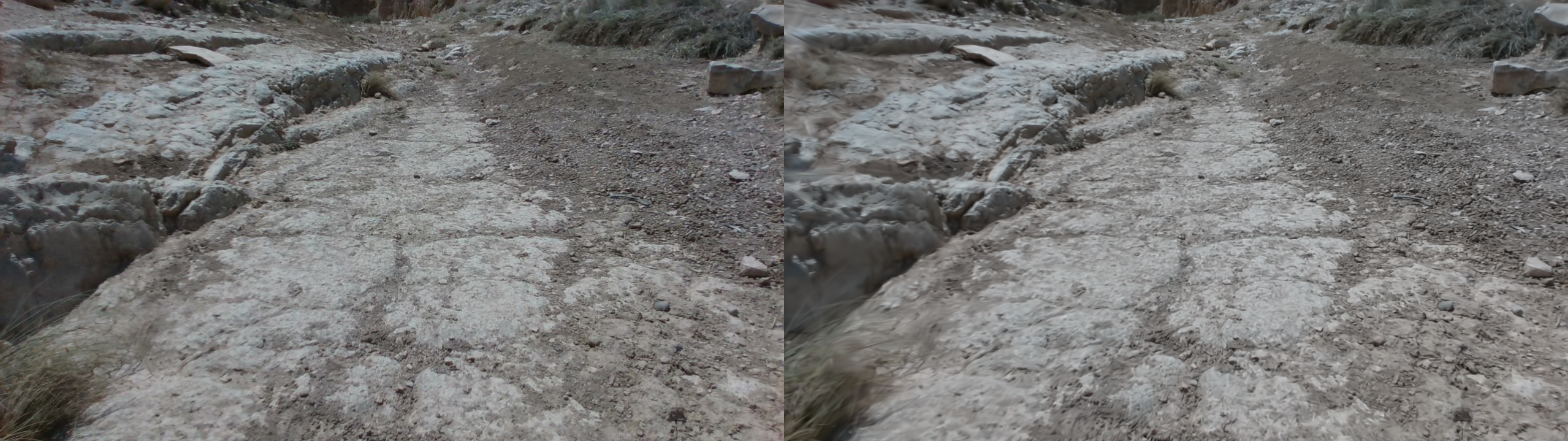}\\[2mm]

    \includegraphics[width=\textwidth]{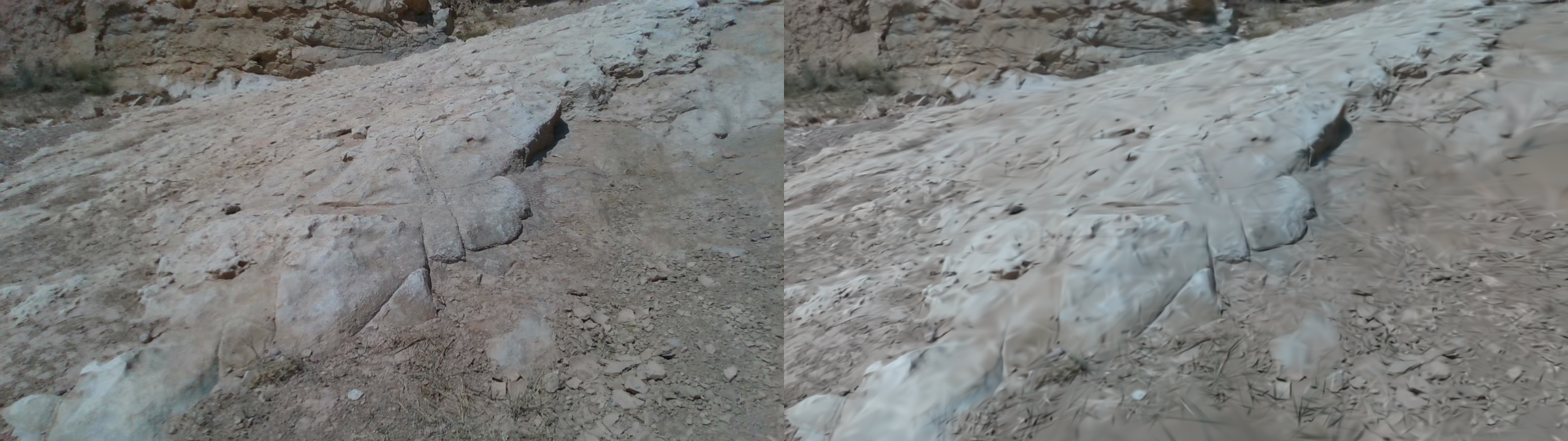}
    \caption{Comparative examples between real-world views (left) and their corresponding 3D reconstructions (right). The visual alignment demonstrates the high geometric and radiometric consistency achieved by the proposed workflow.}
    \label{fig:eval_comparison}
\end{figure*}

Representative examples of the reconstructed scenes are shown in \autoref{fig:aerial_views} and \autoref{fig:eval_comparison}. 
These visual results illustrate the capability of the proposed pipeline to recover coherent 3D structure and radiometric appearance from real-world rover data acquired under uncontrolled outdoor conditions. 
The models correspond to Mars-analog terrains in the Bardenas Reales site, characterized by heterogeneous textures, irregular topography, and variable natural illumination.

The aerial views in \autoref{fig:aerial_views} highlight the overall surface continuity and topographic consistency achieved in the reconstructed models, while \autoref{fig:eval_comparison} compares real camera images with their corresponding renderings. 
The visual alignment between the two domains demonstrates the geometric and photometric consistency attained through the Gaussian Splatting approach implemented in \textit{Splatfacto-W}.

\vspace{3mm}

\subsection{Performance Metrics}

\label{tablasresultados}

To quantitatively evaluate the proposed pipeline, both computational performance and reconstruction quality were analyzed using the metrics outlined in \autoref{tab:comparison_preprocessing} and \autoref{tab:combined_quality}.  
These indicators quantify the efficiency of preprocessing and the geometric and photometric accuracy of the resulting pipeline.

\autoref{tab:comparison_preprocessing} compares the execution times of three preprocessing configurations:  
(1) the default \textit{Nerfstudio} method,  
(2) a sequential image-matching approach optimized for ordered rover trajectories, and  
(3) the proposed configuration, which combines sequential matching with GNSS-based camera priors. In GNSS-denied scenarios, these priors could be replaced by camera pose estimations computed through \ac{VO} or any other available estimation.  
The \textbf{Match [s]} column measures the time required for feature correspondence between images, while \textbf{Map [min]} indicates the duration of the mapping stage, where camera poses and sparse 3D structures are estimated.  
The \textbf{Conv.} column denotes whether the reconstruction converged successfully.  
Together, these values reflect the computational cost and robustness of the geometric preprocessing pipeline.

\autoref{tab:combined_quality} focuses on the quality of the resulting reconstructions.  
The \textbf{Reprojection Error} expresses the average pixel deviation between the reprojected 3D points and their actual image locations (a lower value indicates higher geometric accuracy).  
The \textbf{Avg.\ Obs.} (average number of observations per image) represents how many distinct 3D points are typically visible in each frame, which relates to the redundancy and overlap of the input data.  
\textbf{Track Len.} measures the mean number of images in which each 3D point is observed, assessing temporal coherence across the trajectory.  

To complement the geometric assessment, visual fidelity was measured using three standard rendering metrics:  
\textbf{\acs{PSNR}} (\acl{PSNR}) quantifies pixel-level similarity between rendered and real images;  
\textbf{\acs{SSIM}} (\acl{SSIM}) evaluates perceived structural consistency;  
and \textbf{\acs{LPIPS}} (\acl{LPIPS}) approximates perceptual quality from a human-vision standpoint.  
For these metrics, higher PSNR and \ac{SSIM} values and lower \ac{LPIPS} scores indicate more realistic visual reconstructions.

\begin{table*}[t]
\centering
\caption{Comparison of reconstruction times for different preprocessing configurations: Nerfstudio default, sequential matching, and our proposed method with sequential matching and camera priors.}
\small
\begin{tabularx}{\textwidth}{Y Y Y Y Y Y Y Y Y Y}
\toprule
& \multicolumn{3}{c}{\textbf{Nerfstudio (Default)}} 
& \multicolumn{3}{c}{\textbf{Sequential Matching}} 
& \multicolumn{3}{c}{\textbf{Proposed Method (Seq. + Priors)}} \\
\cmidrule(lr){2-4} \cmidrule(lr){5-7} \cmidrule(lr){8-10}
\textbf{Trajectory} & \textbf{Match [min]} & \textbf{Map [min]} & \textbf{Conv.} 
& \textbf{Match [min]} & \textbf{Map [min]} & \textbf{Conv.} 
& \textbf{Match [min]} & \textbf{Map [min]} & \textbf{Conv.} \\
\midrule
T1 & 6.18 & 68.1  & \cellcolor{excelente}Yes & 0.447 & 56.37 & \cellcolor{excelente}\cellcolor{excelente}Yes & 0.441 & 26.858 & \cellcolor{excelente}\cellcolor{excelente}Yes \\
T2 & 5.32 & 50.3  & \cellcolor{deficiente}No  & 0.353 & 25.00 & \cellcolor{excelente}\cellcolor{excelente}Yes & 0.354 & 19.354 & \cellcolor{excelente}\cellcolor{excelente}Yes \\
T3 & 13.8 & 143.85 & \cellcolor{excelente}\cellcolor{excelente}Yes & 1.2 & 54.564 & \cellcolor{excelente}\cellcolor{excelente}Yes & 1.2 & 51.00 & \cellcolor{excelente}\cellcolor{excelente}Yes \\
T4 & 6.53 & 96.57 & \cellcolor{deficiente}No  & 0.366 & 51.197 & \cellcolor{deficiente}No  & 0.36 & 22.995 & \cellcolor{excelente}\cellcolor{excelente}Yes \\
T5 & 6.25 & 73.6 & \cellcolor{deficiente}No   & 0.297 & 40.88 & \cellcolor{excelente}\cellcolor{excelente}Yes & 0.319 & 25.725 & \cellcolor{excelente}Yes \\
\bottomrule
\end{tabularx}
\label{tab:comparison_preprocessing}
\end{table*}

\begin{table*}[h!]
\centering
\caption{Integrated evaluation of reconstruction quality across all trajectories. Color coding follows the legend below, emphasizing higher-quality results in green tones.}
\small
\setlength{\tabcolsep}{5pt}
%\vspace{2cm}
\begin{tabularx}{\textwidth}{Y Y Y Y Y Y Y}
\toprule
\textbf{Trajectory} &
\textbf{Reproj. Err. [px]} &
\textbf{Avg. Obs. [\#]} &
\textbf{Track Len. [\#]} &
\textbf{PSNR [dB]} &
\textbf{SSIM} &
\textbf{LPIPS} \\
\midrule
T1 & \cellcolor{aceptable}0.552 & \cellcolor{excelente}6437.78 & \cellcolor{excelente}7.07 & \cellcolor{buena}28.2 $\pm$ 3.0 & \cellcolor{buena}0.806 $\pm$ 0.116 & \cellcolor{buena}0.231 $\pm$ 0.119 \\
T2 & \cellcolor{aceptable}0.548 & \cellcolor{excelente}6287.21 & \cellcolor{buena}6.98 & \cellcolor{aceptable}27.1 $\pm$ 1.8 & \cellcolor{buena}0.803 $\pm$ 0.048 & \cellcolor{buena}0.204 $\pm$ 0.041 \\
T3 & \cellcolor{aceptable}0.541 & \cellcolor{buena}5594.43 & \cellcolor{excelente}7.43 & \cellcolor{buena}28.2 $\pm$ 3.0 & \cellcolor{buena}0.805 $\pm$ 0.116 & \cellcolor{buena}0.231 $\pm$ 0.119 \\
T4 & \cellcolor{aceptable}0.520 & \cellcolor{excelente}6141.99 & \cellcolor{buena}6.91 & \cellcolor{aceptable}26.1 $\pm$ 3.5 & \cellcolor{aceptable}0.783 $\pm$ 0.107 & \cellcolor{aceptable}0.264 $\pm$ 0.121 \\
T5 & \cellcolor{buena}0.476 & \cellcolor{buena}5310.43 & \cellcolor{excelente}7.92 & \cellcolor{buena}29.3 $\pm$ 3.9 & \cellcolor{buena}0.827 $\pm$ 0.133 & \cellcolor{buena}0.224 $\pm$ 0.101 \\
\midrule
\cellcolor{excelente}\textbf{Excellent} & \qty{<0.2}{px}          & $>6000$    & $>7$ & \qty{>30}{dB}         & $>0.9$ & $<0.15$ \\
\cellcolor{buena}\textbf{Good}          & \qtyrange{0.2}{0.5}{px} & 4000--6000 & 5--7 & \qtyrange{28}{30}{dB} & 0.8--0.9 & 0.15--0.25 \\
\cellcolor{aceptable}\textbf{Acceptable}& \qtyrange{0.5}{1.0}{px} & 2000--4000 & 3--5 & \qtyrange{25}{28}{dB} & 0.7--0.8 & 0.25--0.35 \\
\cellcolor{deficiente}\textbf{Poor}     & \qty{>1.0}{px}          & $<2000$    & $<3$ & \qty{<25}{dB}         & $<0.7$ & $>0.35$ \\
\bottomrule
\end{tabularx}
\label{tab:combined_quality}
\end{table*}

\subsection{Results Evaluation and Limitations}

Overall, the developed workflow successfully reconstructed all trajectories with consistent geometric and visual quality. Three algorithmic configurations were compared: (1) the default \textit{Nerfstudio} preprocessing, (2) a sequential matching implementation, and (3) our proposed method combining sequential matching with GNSS-based priors. The latter achieved the best results, reducing average matching and mapping times by \qty{93.1}{\%} and \qty{65.8}{\%}, respectively, while maintaining full reconstruction convergence. This demonstrates the benefits of integrating sequential matching and prior motion information when dealing with ordered, trajectory-based image datasets.

From the geometric perspective, the \ac{SfM} evaluation yielded low mean reprojection errors (below \qty{0.6}{px}), high average observations per image (above 5,000), and track lengths exceeding 7 frames in most cases, confirming the internal consistency and robustness of COLMAP’s reconstruction. These results indicate sufficient overlap and visual redundancy, even in scenes with limited texture.

Quantitative evaluation of the radiance field rendering was performed using \ac{PSNR}, \ac{SSIM}, and \ac{LPIPS} metrics. Across the five trajectories, average \ac{PSNR} values ranged between 26 and \qty{29}{dB}, \ac{SSIM} between 0.78 and 0.83, and \ac{LPIPS} between 0.20 and 0.26, denoting acceptable-to-good perceptual quality. Among all the trajectories evaluated, T5 achieved the best results. This appears to be due to its highly regular and almost rectilinear capture pattern, which improves feature tracking and results in a more stable and well-conditioned bundle adjustment. However, a limited viewpoint diversity reduces the geometric coverage of the scene, worsening the quality of the reconstruction when observed from previously unseen angles. Despite the use of training views for evaluation, which potentially inflates absolute scores, the results consistently reflect coherent color reproduction, geometric alignment, and visually plausible renderings.

A qualitative inspection of the reconstructions corroborates these findings: reconstructed models exhibit accurate geometry and continuous surfaces, though thin objects and vegetation remain challenging. Some regions near the image boundaries exhibit ghosting and aliasing artifacts, typical of low-overlap sequences. Nevertheless, the temporal coherence between consecutive rendered frames is satisfactory, which is essential for navigation and visualization applications.

In summary, the proposed pipeline achieved reliable 3D reconstructions under Mars-analog conditions, balancing visual fidelity, computational efficiency, and robustness. Experiments were conducted on five representative segments from the BASEPROD dataset, all acquired at a single Mars-analog site. Although these segments exhibit variability in terrain texture, illumination conditions, and rover trajectory geometry, cross-site validation on additional planetary-analog datasets is still required to fully assess generalization. Nevertheless, the current results confirm the viability of the approach for rover-based scene reconstruction and perception.
However, several limitations of the current workflow must be acknowledged. Despite its robustness, the pipeline still relies on offline processing and requires computational resources far beyond those available on present-day planetary rover platforms, preventing real-time or onboard deployment. Performance also depends heavily on COLMAP’s SfM stage, which constitutes the main bottleneck of the workflow: although the results reported above demonstrate good average consistency, sparse reconstruction remains sensitive to texture richness, viewpoint overlap, and lighting variability. Any degradation in COLMAP’s initialization or feature matching propagates downstream and directly affects reconstruction quality.
Importantly, Gaussian Splatting is particularly vulnerable to limited viewpoint diversity. Predominantly frontal or straight-line trajectories often provide insufficient parallax for stable splat optimization, reducing the model’s ability to generalize to novel viewpoints even when sparse geometry is nominally acceptable. Another practical limitation may arise from inherent ROS2 image acquisition pipeline irregularities as inconsistent frame rates or timestamping reduce the effectiveness of sequential matching, negatively affecting both SfM and radiance field optimization. 

\section{CONCLUSION}
\label{sec:conclusions}

This work presented a fully automated and modular pipeline for 3D reconstruction from rover data, integrating a version of \textit{COLMAP} with prior-based structure estimation and dense reconstruction using Gaussian Splatting within \textit{Nerfstudio}. The pipeline is designed to handle \ac{ROS2}-based datasets, standardize preprocessing, and produce high-quality photorealistic models, while providing quantitative metrics for both geometric and visual evaluation. We validated our proposed workflow on real rover data from the \ac{BASEPROD} dataset, demonstrating reliable reconstructions across trajectories. Quantitative results show that incorporating sequential matching and pose priors reduces matching and mapping times by \qty{93.1}{\%} and \qty{65.8}{\%}, respectively, compared to Nerfstudio's default configuration, while improving reconstruction accuracy and convergence. 

While our evaluation focused on preprocessing configurations and internal pipeline optimization, a direct end-to-end comparison against alternative mapping approaches (e.g., SLAM) and radiance-field implementations remains an important direction for future studies. %This, together with extended ablation studies, would further validate the selection of Splatfactor-W under domain-specific constraints. 

Although the modified process achieved successful reconstructions in Mars-analog environments, it still faces limitations in speed, robustness, and data dependency. The pipeline currently operates offline, preventing real-time mapping or onboard decision-making, and relies on COLMAP for structure estimation, which makes performance sensitive to initialization, texture richness, viewpoint overlap, and lighting variability. These factors can lead to inconsistent results, especially in low-texture or repetitive terrains, and should, therefore, guide future improvements. %Additionally, a more granular analysis of computational efficiency and scalability on larger or longer-duration datasets, would provide additional insights into deployment feasibility. 

Overall, this work confirms the feasibility and potential of these technologies, providing a documented and operational foundation for future research in radiance-field-based spatial reconstruction using rover platforms.

%Increasing efficiency and robustness through improved data acquisition and preprocessing are a must. expanding viewpoint coverage with 360° or multi-camera setups, and integrating real-time localization modules such as LiDAR-based or visual–inertial SLAM. Additionally, adaptive filtering and uncertainty estimation will be explored to enhance the reliability of the reconstructed models.
% In applications with greater communication and computing capacity, the developed tools represent a robust solution for 3D mapping and reconstruction tasks. 

\section*{ACKNOWLEDGMENTS}

This work was part of the project INSIGHT (PID2024-160373OB-C21) funded by MICIU / AEI / 10.13039/501100011033 / FEDER, UE. It was partially supported by the European Space Agency under activity no.\ 4000140043/22/NL/GLC/ces. It was also technically supported by IMECH.UMA through PPRO-IUI-2023-02.

\bibliographystyle{IEEEtran}
\bibliography{references}

\vfill
\end{document}